\documentclass[fleqn,10pt]{wlscirep}
\usepackage[utf8]{inputenc}
\usepackage[T1]{fontenc}
\usepackage{graphicx}
\usepackage{subcaption}
\usepackage{caption}
\usepackage{booktabs}
\usepackage[dvipsnames]{xcolor}
\usepackage[table,xcdraw]{xcolor}
\usepackage{comment}
\usepackage{amsmath} 
\definecolor{YellowGreen}{HTML}{D9EAD3} % light green (same as in plot)
\definecolor{Peach}{HTML}{F4CCCC}       % light red (same as in plot)
\definecolor{SkyBlue}{HTML}{CFE2F3}     % light blue for question box
\usepackage{times}
\newcommand{\commentout}[1]{}
\usepackage{latexsym}
\usepackage{adjustbox}
\usepackage{multirow}
\usepackage{longtable}
\usepackage{booktabs}

\title{HealthContradict: Evaluating Biomedical Knowledge Conflicts in Language Models}

\author[1*]{Boya Zhang}
\author[1]{Alban Bornet}
\author[2]{Rui Yang}
\author[3,4]{Nan Liu}
\author[1]{Douglas Teodoro}

\affil[1]{Faculty of Medicine, University of Geneva, Geneva, Switzerland}
\affil[2]{Department of Biomedical Informatics, Yong Loo Lin School of Medicine, National University of Singapore}
\affil[3]{Department of Biostatistics \& Bioinformatics, Duke University}
\affil[4]{Artificial Intelligence Institute, National University of Singapore}

\affil[*]{boya.zhang@unige.ch}

\begin{abstract}
How do language models use contextual information to answer health questions? How are their responses impacted by conflicting contexts? We assess the ability of language models to reason over long, conflicting biomedical contexts using \textsc{HealthContradict}, an expert-verified dataset comprising 920 unique instances, each consisting of a health-related question, a factual answer supported by scientific evidence, and two documents presenting contradictory stances. We consider several prompt settings, including correct, incorrect or contradictory context, and measure their impact on model outputs. 
Compared to existing medical question-answering evaluation benchmarks, \textsc{HealthContradict} provides greater distinctions of language models' contextual reasoning capabilities. Our experiments show that the strength of fine-tuned biomedical language models lies not only in their parametric knowledge from pretraining, but also in their ability to exploit correct context while resisting incorrect context.
\end{abstract}
\begin{document}

\flushbottom
\maketitle
\thispagestyle{empty}

\section{Introduction}

Language models are susceptible to generating reasonable yet nonfactual content \cite{10.1145/3703155}. This issue raises concerns about the reliability of language models in providing medical advice, as there are significant risks when they generate convincing but incorrect information, which could influence people's health-related decisions \cite{10.1145/3121050.3121074}. Additionally, knowledge and misinformation in the biomedical domain both evolve rapidly, especially during medical crises \cite{doi:10.1073/pnas.1710755115}, with unverified information spreading quickly across the internet \cite{wang2019systematic},  impacting pre-training and in-context learning for these models.

\begin{figure}[!h]
  \centering
  \includegraphics[scale=0.43]{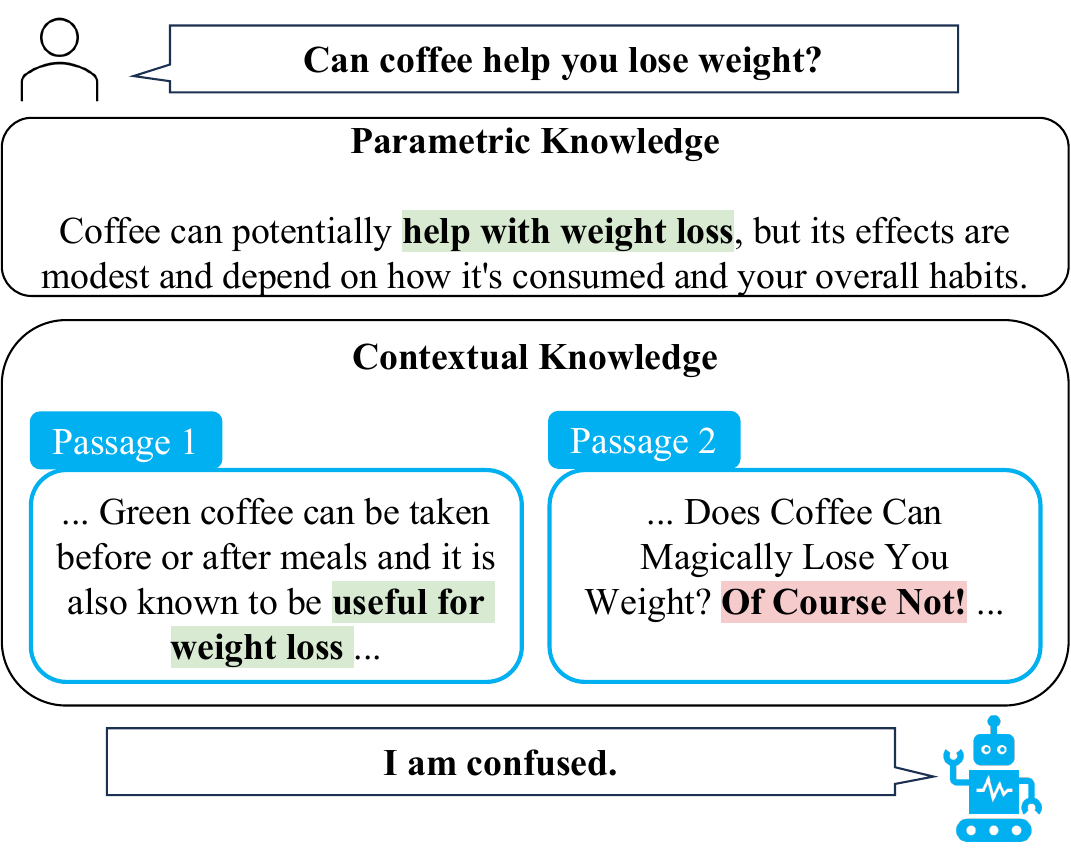}
  \caption{Biomedical Knowledge Conflicts Make Language Models Confused.}
  \label{fig:kg}
\end{figure}

Existing methods to mitigate misinformation use static fact sources for hallucination detection \cite{farquhar2024detecting, li2024dawn} or verified evidence to refute false claims \cite{thorne-etal-2018-fever, wadden-etal-2020-fact, 10.1145/3561389, vladika-etal-2024-healthfc}. 
These strategies have often been combined in retrieval-augmented generation (RAG) \cite{lewis2020retrieval} pipelines, as one of the most effective methods to attenuate hallucinations in the biomedical domain \cite{he2025retrieval}. 
Despite some attempts considering information quality \cite{ZHANG2024}, current approaches for biomedical RAG \cite{10.1093/bioinformatics/btae238, LI2025104769, yang2025retrieval} primarily focus on improving relevance in the retrieval pipeline \cite{jin2021disease,pmlr-v174-pal22a, taboureau2010chemprot, segura2013semeval, gurulingappa2012development}.
However, in the real world, contradictory sources could be used to verify the same claim, leading to knowledge conflicts \cite{xu-etal-2024-knowledge-conflicts} in RAG paradigms. 
For instance, consider the situation illustrated in  Figure~\ref{fig:kg}, where a language model has its own parametric knowledge, i.e., learned during pre-training, stating that coffee aids in weight loss. 
However, when utilizing an in-context learning approach, the model is presented with two contradictory passages as contextual knowledge, i.e., information from the external source material, while answering the question. In this case, the conflicts arise from the contradictions between Passage 1 and Passage 2, as well as between the model's parametric knowledge and Passage 2. 

The behavior of language models is influenced by knowledge conflicts \cite{xu-etal-2024-knowledge-conflicts}, which are the contradictions within parametric knowledge learned at training time \cite{singhal2023large} and contextual knowledge given at inference time \cite{10.1145/3560815, info:doi/10.2196/60501, 10.1145/3637528.3671470, shi-etal-2024-replug}. Language models are receptive to coherent contextual knowledge when it conflicts with parametric knowledge \cite{xie2024adaptive}. Multi-turn persuasive conversations as contextual knowledge can even manipulate language models’ factual parametric knowledge  \cite{xu-etal-2024-earth}. On the other hand, language models are biased toward parametric knowledge when the contextual knowledge is self-contradictory \cite{jin-etal-2024-tug, xie2024adaptive}. They also have difficulty generating answers that reflect the self-contradiction of the contextual knowledge \cite{chen-etal-2022-rich}, especially for implicit conflicts that require reasoning \cite{hou2024wikicontradict}. Besides, language models struggle with self-contradictions in long documents that require more nuance and context \cite{li-etal-2024-contradoc}.

Various context-aware methods were proposed to overcome language models' confusion regarding knowledge conflicts. While context-aware decoding overrode a model's parametric knowledge when it contradicts the contextual knowledge \cite{shi2024trusting}, ContextCite traced back the parts of the contextual knowledge that led a model to generate a particular statement to improve the explainability of language models \cite{cohen2024contextcite}. However, these methods only focused on either parametric or contextual knowledge. COMBO \cite{zhang-etal-2023-merging}, on the other hand, leveraged both the parametric and contextual knowledge by using discriminators trained on silver labels to assess passage compatibility. In addition, DisentQA \cite{neeman-etal-2023-disentqa} trained a model that predicts two types of answers, one based on contextual knowledge and one on parametric knowledge for a given question. Contrastive decoding further maximized the difference between logits under knowledge conflicts and calibrates the model's probability in the correct answer \cite{jin-etal-2024-tug}. Solutions were also proposed to mitigate the harmful behavior of language models. At the training phase, counterfactual and irrelevant contexts were injected into standard supervised datasets to perform knowledge-aware finetuning to enhance language models' robustness \cite{li-etal-2023-large}. Meanwhile, in-context pretraining enhanced language models' performance in complex contextual reasoning \cite{shi2024context}. At the inference phase, defense strategies of misinformation detection, vigilant prompting, and reader ensembles were proposed to mitigate misinformation generated by language models \cite{pan2023risk}. In addition, query augmentation was used to search for robust answers to defend against poisoning attacks \cite{weller-etal-2024-defending}. Furthermore, fact duration prediction identified which facts are prone to rapid change and helps models avoid reciting outdated information \cite{zhang2023mitigating}. Current approaches prioritized mitigating either contextual conflicts or harmful behaviors of language models. However, both context-awareness and truthfulness are important in improving the answers of language models in the biomedical domain.

Language models' behavior on general domain knowledge conflicts have been evaluated with synthetic datasets featuring explicit and simple contradictions \cite{longpre-etal-2021-entity, chen-etal-2022-rich, wang2024resolving, li-etal-2024-contradoc, wu2024clasheval}, as well as real-world datasets featuring implicit and complex contradictions \cite{9671319, hou2024wikicontradict}. While research on knowledge conflicts primarily focuses on general domains, its impact on the biomedical domain remain underexplored. Conflicts in biomedical knowledge are complex due to the domain's distinctive lexicon and the complex syntax of long sentences \cite{ondov2022survey}. ManConCorpus \cite{alamri2016corpus} collected contradictory claims from biomedical literature addressing 24 cardiovascular research questions. Meanwhile, COVID-19 NLI \cite{sosa-etal-2023-detecting} automatically identified contradictory claims about COVID-19 drug efficacy from the subset of CORD-19 \cite{wang-etal-2020-cord}. In addition, ClashEval  \cite{wu2024clasheval} sampled drug information pages from UpToDate.com and modified the numerical drug dosages with GPT-4o to create contradictions. On a broader range of medical topics, MedNLI \cite{romanov-shivade-2018-lessons} was manually curated by creating contradicting, entailing, and neutral sentences paired with clinical descriptions. In contrast to manually curated sentences, \cite{makhervaks-etal-2023-clinical} focused on identifying naturally occurring sentences containing clinical outcomes and detecting potential contradictions using the SNOMED-CT ontology \cite{stearns2001snomed}. These datasets aim to identify contradictions in biomedical sentences but lack evidence to determine the correct claims. Furthermore, systems to identify sentence-level contradictions are not helpful when contradictions are conveyed across multiple sentences in longer texts.

To address these limitations, we propose \textsc{HealthContradict}, a dataset consisting of 920 unique instances, each comprising a health-related question and two documents with contradictory stances. In addition, each instance has a factual answer supported by scientific evidence. Using \textsc{HealthContradict}, we evaluated several language models, from 1B to 8B parameters, including general domain and its biomedical counterpart, to answer health-related questions in the presence of knowledge conflicts. When provided with biomedical context and an language model, our benchmark evaluates: (i) How do language models answer biomedical questions in the presence of knowledge conflicts? (ii) How does the biomedical context provided to the language models act as a causal factor in inducing the answer? To do so, we include correct, incorrect or contradictory context in different prompt scenarios, and assess models' accuracy and probability distribution in answering health-related questions.

Our contributions are the following: (i) A novel dataset - \textsc{HealthContradict} - designed to evaluate language models in presence of conflicting information in the biomedical domain; (ii) We perform a comprehensive evaluation of language models against \textsc{HealthContradict}, assessing their ability to reason over long, conflicting biomedical contexts using interpretable quantitative metrics; and (iii) Moreover, we compare general domain language models vs. their fine-tuned biomedical counterparts and reveal that the strength of the latter lies in their ability to exploit correct while resisting incorrect contextual knowledge.

\section{Results}

\subsection{\textsc{HealthContradict} Benchmark}

We created the \textsc{HealthContradict} benchmark, a dataset of 920  instances. Each instance is a health-related question with a factual answer supported by scientific evidence, paired with two documents presenting contradictory stances. Each document appears only once in the entire dataset to ensure unbiased evaluation. Table~\ref{tab:example-instance} shows an example, where for a given health-related question (\textit{``Can coffee help you lose weight?''}), two contradictory documents are provided (\texttt{yes}: \textit{``... useful for weight loss ...''} vs. \texttt{no}: \textit{``... Of Course Not! ...''}), together with the factual answer (\texttt{yes}) supported by scientific evidence \textit{``... caffeine intake might promote weight, BMI and body fat reduction ...''}).

\begin{figure}[!h]
    \centering
    \includegraphics[width=0.5\linewidth]{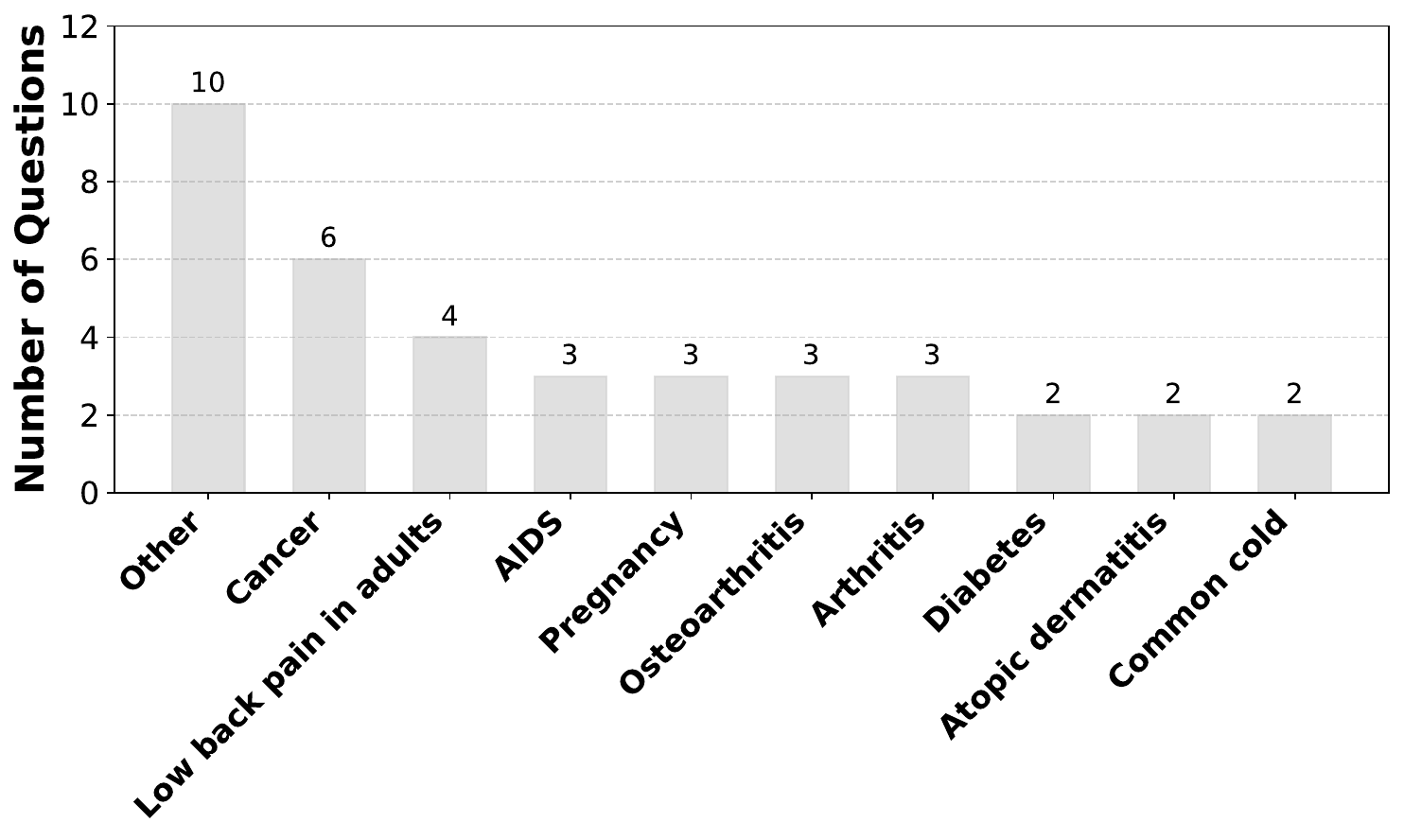}
    \caption{Example disease and condition categories in \textsc{HealthContradict}}
    \label{fig:tdd}
\end{figure}

In total, the dataset contains 81 questions, each addressing a health issue and a potential treatment. These issues span 50 disease and condition categories, such as “Cancer”, “Low back pain in adults”, and “AIDS”, as well as one general well-being category “Other,” which covers topics such as weight management. Figure \ref{fig:tdd} illustrates 10 example disease and condition categories most commonly addressed by the health questions. The complete list of all questions and their corresponding categories is provided in the Appendix Table~\ref{tab:full-question-table}.

The questions are associated with 1,840 documents (920 with stance \texttt{yes} and 920 with stance \texttt{no}). The factual answers are "yes" for 26.5\% and "no" for 73.5\% of the instances. We chose not to balance the dataset, as real-world scenarios are often imbalanced, and our goal is to evaluate models under practical use cases. The average document length is 2,347 words, varying from 23 to 30,444. See Appendix Table~\ref{tab:dataset-stats} and Table~\ref{tab:top-domains} for further information on the dataset. 

\begin{table*}[!ht]
    \centering
    
    \begin{tabular}{p{3.5cm} p{10cm} p{1cm}}
    \toprule
    \textbf{Field} & \textbf{Content} & \textbf{Stance}\\
    \midrule
    \textbf{Question} & Can coffee help you lose weight? & \\
    \textbf{Correct Document} & ... Green coffee can be taken before or after meals and it is also known to be \colorbox{YellowGreen}{useful for weight loss} ... & \colorbox{YellowGreen}{Yes}\\
    \textbf{Incorrect Document} & ... Does Coffee Can Magically Lose You Weight? \colorbox{Peach}{Of Course Not!} ... & \colorbox{Peach}{No} \\
    \textbf{Scientific Evidence} & ... Overall, the current meta-analysis demonstrated that \colorbox{YellowGreen}{caffeine intake might promote weight, BMI and body fat reduction}. & \colorbox{YellowGreen}{Yes}\\
    \bottomrule
    \end{tabular}
    \caption{An Example Instance from \textsc{HealthContradict} Dataset. Each Instance includes a Health Question, Two Contradictory Documents, the Factual Answer, and Supporting Scientific Evidence.}
    \label{tab:example-instance}
\end{table*}

%\subsection{Prompt Template}

\begin{table}[!h]
\centering
\begin{tabular}{@{}c p{15cm}@{}}
\toprule
 & \textbf{Prompt Template} \\
\midrule
\textbf{NC} & \textbf{Instruction:} Answer the following question with only \texttt{YES} or \texttt{NO} based on your parametric knowledge. \newline
\textbf{Question:} \{\colorbox{SkyBlue}{Question}\} \\
\addlinespace
\textbf{CC} & \textbf{Instruction:} Answer the following question with only \texttt{YES} or \texttt{NO} based on the given contextual knowledge. \newline
\textbf{Question:} \{\colorbox{SkyBlue}{Question}\} \newline
\textbf{Context:} \{\colorbox{YellowGreen}{Correct Document}\} \\
\addlinespace
\textbf{IC} & \textbf{Instruction:} Answer the following question with only \texttt{YES} or \texttt{NO} based on the given contextual knowledge. \newline
\textbf{Question:} \{\colorbox{SkyBlue}{Question}\} \newline
\textbf{Context:}  \{\colorbox{Peach}{Incorrect Document}\} \\
\addlinespace
\textbf{CIC} & \textbf{Instruction:} Answer the following question with only \texttt{YES} or \texttt{NO} based on the given contextual knowledge. \newline
\textbf{Question:} \{\colorbox{SkyBlue}{Question}\} \newline
\textbf{Context:} \{\colorbox{YellowGreen}{Correct Document}\} \newline\hspace*{4em}\{\colorbox{Peach}{Incorrect Document}\} \\
\addlinespace
\textbf{ICC} & \textbf{Instruction:} Answer the following question with only \texttt{YES} or \texttt{NO} based on the given contextual knowledge. \newline
\textbf{Question:} \{\colorbox{SkyBlue}{Question}\} \newline
\textbf{Context:} \{\colorbox{Peach}{Incorrect Document}\} \newline\hspace*{4em}\{\colorbox{YellowGreen}{Correct Document}\}\\
\bottomrule
\end{tabular}
\caption{Prompt templates for contextual evaluation in \textsc{HealthContradict}. 
\textbf{NC}: No Context, 
\textbf{CC}: Correct Context, 
\textbf{IC}: Incorrect Context, 
\textbf{CIC}: Correct + Incorrect Context, 
\textbf{ICC}: Incorrect + Correct Context.}
\label{tab:prompt-templates}    
\end{table}

To investigate how language models respond to real-world biomedical knowledge conflicts, we develop five prompt templates to evaluate their performance under different question-answering scenarios. As illustrated in Table~\ref{tab:prompt-templates}, for each annotated instance from the \textsc{HealthContradict} dataset, we generate five question prompts based on these pre-defined templates. These prompts were presented to each model independently, and the model did not retain any state or output from other prompts. The controlled set of prompt templates enable us to evaluate the effect of correct, incorrect, or conflicting context without introducing additional variability in the phrasing of the prompt. We note that this design choice is consistent with prior work such as WikiContradict \cite{HouEtAl2024_WikiContradict}, which also employed minimal instructional phrasing to perform comparative analysis. Specifically, Prompt NC evaluates models' parametric knowledge (control template), while Prompt CC and IC examine their performance with a single document provided as context (correct and incorrect, respectively). Prompt CIC and ICC, on the other hand, assess a model's ability to handle health questions in the presence of a conflicting contextual information. The difference between Prompt CIC and ICC aims to evaluate if the position of the contradictory document influences the model's answer. These include one without context (Prompt NC) and four with varying context configurations. For each of the 920 instances, we applied 4 context-based prompt templates. Additionally, we include 81 prompts for each question from the Prompt NC, leading to a total of 3,761 prompts used for comparative evaluation. We focused on yes/no questions to enable controlled analysis of model behavior under conflicting contexts, as the binary classification offers a clear setting for evaluation in this scenario. 

\subsection{Baseline Models}

Our baseline selection was motivated by two objectives: (i) to evaluate whether biomedical fine-tuning enhances performance within the biomedical domain and (ii) to assess whether increasing model size leads to performance gains. We also selected language models with extended context lengths to process long documents. We show details of selected models in Table~\ref{tab:model-category-param}. Each biomedical model was obtained by fine-tuning the general-domain model presented in the same row. \textit{Instruct} refers to models that have been finetuned to follow user instructions (i.e., instruction‑tuned). We do not finetune any models and perform zero-shot inference for the selected baseline models. We evaluate open-source language models ranging from 1B to 8B parameters for reproducibility of resource-limited healthcare settings.

\subsection{Baseline Benchmarks}

We compare our benchmark to three widely used multiple-choice question answering benchmarks. MedMCQA~\cite{pmlr-v174-pal22a}, MedQA-4-Option~\cite{jin2021disease} and PubMedQA~\cite{jin2019pubmedqa}. MedMCQA~\cite{pmlr-v174-pal22a} and MedQA-4-Option~\cite{jin2021disease} are derived from medical exam questions and evaluate the model performance on clinical medical knowledge. PubMedQA~\cite{jin2019pubmedqa} is derived from PubMed~\cite{pubmed} articles and evaluates the model performance on theoretical medical knowledge. 

We evaluate selected baseline models using Language Model Evaluation Harness \cite{eval-harness}. All results are reported using accuracy, which measures the proportion of questions answered correctly. For MedMCQA~\cite{pmlr-v174-pal22a}, we use the  validation split which contains 4,183 four-option multiple-choice questions. For MedQA-4-Option, we evaluate on the test split, comprising 1,273 four-option multiple-choice questions. For PubMedQA~\cite{jin2019pubmedqa}, we use test split, which includes 500 three-option multiple-choice questions.

As shown in Table~\ref{tab:medical-benchmarks}, the difference among larger language models (7-8B) is minor. Moreover, fine-tuned biomedical model \textsc{Meditron3‑Qwen2.5‑7B} underperforms \textsc{Qwen2.5‑7B}. These state-of-the-art evaluation benchmarks are weak at comprehending difference among models' capabilities because they primarily assess a model’s parametric knowledge. 

\begin{table}[!h]
\centering
\begin{tabular}{lrrr}
\toprule
\multicolumn{2}{c}{\textbf{Domain}} & \textbf{Size} & \textbf{Context Length} \\
\cmidrule(lr){1-2}
\textbf{General} & \textbf{Biomedical} & & \\
\midrule
Llama-3.2-1B-Instruct \cite{meta-llama-3.2-1b-instruct}     & BioMed-Llama-3.2-1B \cite{~bio-medical-1b-cot} & 1B  & 128K \\
Qwen2.5-7B \cite{qwen-2.5-7b}               & Meditron3-Qwen2.5-7B \cite{meditron3-qwen-2.5-7b}         & 7B  & 131K \\
Llama-3.1-8B-Instruct \cite{meta-llama-3.1-8b-instruct}    & Meditron3-8B \cite{meditron3-8b} & 8B  & 128K \\
\bottomrule
\end{tabular}
\caption{Domains, Parameter Sizes and Context Lengths of Selected Language Models.}
\label{tab:model-category-param}
\end{table}

\begin{table*}[!ht]
\rowcolors{2}{white}{gray!15}
\centering
\begin{tabular}{lccc}
\toprule
\textbf{Model} & \textbf{MedMCQA} & \textbf{MedQA} & \textbf{PubMedQA} \\
\midrule
Llama-3.2-1B-Instruct \cite{meta-llama-3.2-1b-instruct} & 41.33 & 39.59 & 60.20 \\
BioMed-Llama-3.2-1B \cite{~bio-medical-1b-cot} & 34.88 & 37.39 & 60.40 \\
Qwen2.5-7B \cite{qwen-2.5-7b} & \textbf{60.10} & \textbf{64.49} & 75.20 \\
Meditron3-Qwen2.5-7B \cite{meditron3-qwen-2.5-7b}      & 57.14 & 61.82 & 74.40 \\
Llama-3.1-8B-Instruct \cite{meta-llama-3.1-8b-instruct} & 56.99 & 60.25 & 74.20 \\
Meditron3-8B \cite{meditron3-8b} & 57.83 & 63.00 & \textbf{76.80} \\
\bottomrule
\end{tabular}
\caption{Existing Medical QA benchmarks show limited discriminative power across language models.}
\label{tab:medical-benchmarks}
\end{table*}

\subsection{Evaluations on \textsc{HealthContradict}}

We assess how language models use biomedical contextual knowledge to provide a complementary view of their actual performance. We report Accuracy and Macro F1 as evaluation metrics (definitions in Eq.~\ref{acc},~\ref{f1}, and~\ref{mf1}). Accuracy is a simple measurement of the correctness of predictions, while Macro F1 offers a balanced evaluation by equally weighting the performance of each class.

As shown in Table~\ref{tab:template5_comparison}, when the correct context is provided, i.e., Prompt CC, \textsc{Meditron3-8B} achieves the highest accuracy (91.1\%) in the \textsc{HealthContradict} benchmark, outperforming its parametric knowledge (Prompt NC - control) by 8.7 percentage points ($p < 0.001$). 
The second-best performing model, \textsc{Meditron3-Qwen2.5-7B}, achieves an accuracy of 87.6\%, shows an improvement of 3.8 percentage points adding correct context (Prompt CC) over using only its parametric knowledge (Prompt NC - control) ($p < 0.001$). 
As expected, the worst-performing scenario for all the models is when only an incorrect context is provided (Prompt IC). For example, the correct parametric knowledge of \textsc{Meditron3-8B} is overridden by the incorrect context (i.e., a reduction in performance of the control by 21.6 percentage points, $p < 0.001$). 
Interestingly, when conflicting context is provided (Prompt CIC and ICC), all models drop performance compared to the parametric setting (Prompt NC). This effect is less pronounced in \textsc{Meditron3-8B}, with a drop in accuracy between 2.6 percentage points (Prompt ICC, $p < 0.001$) and 2.8 percentage points (Prompt CIC, $p < 0.001$).

\begin{table*}[!h]
\centering
\rowcolors{3}{gray!15}{white}
\begin{tabular}{lcccccccccc}
\toprule
\multirow{2}{*}{\textbf{Model}} & \multicolumn{2}{c}{\textbf{Prompt NC}} & \multicolumn{2}{c}{\textbf{Prompt CC}} & \multicolumn{2}{c}{\textbf{Prompt IC}} & \multicolumn{2}{c}{\textbf{Prompt CIC}} & \multicolumn{2}{c}{\textbf{Prompt ICC}} \\
 & \textbf{Acc.} & \textbf{F1} & \textbf{Acc.} & \textbf{F1} & \textbf{Acc.} & \textbf{F1} & \textbf{Acc.} & \textbf{F1} & \textbf{Acc.} & \textbf{F1} \\
\midrule
Llama-3.2-1B-Instruct \cite{meta-llama-3.2-1b-instruct} & 38.3 & 36.9 & \underline{28.6} & \underline{24.1} & \underline{26.0} & \underline{20.8} & \underline{27.0} & \underline{22.1} & \underline{27.6} & \underline{22.9} \\
BioMed-Llama-3.2-1B \cite{~bio-medical-1b-cot}        & \underline{33.8} & \underline{31.3} & 48.3 & 48.2 & 35.2 & 35.1 & 51.4 & 48.6 & 57.3 & 54.8 \\
Qwen2.5-7B \cite{qwen-2.5-7b}                           & 83.3 & 77.2 & 76.7 & 73.0 & 54.7 & 52.0 & 66.0 & 62.1 & 68.5 & 64.3 \\
Meditron3-Qwen2.5-7B \cite{meditron3-qwen-2.5-7b}       & \textbf{83.8}$^{\dagger}$ & \textbf{78.9}$^{\dagger}$ & 87.6 & 85.0 & 41.5 & 39.9 & 70.2 & 67.7 & 63.2 & 60.2 \\
Llama-3.1-8B-Instruct \cite{meta-llama-3.1-8b-instruct} & 77.9 & 76.5 & 70.0 & 69.1 & 37.8 & 37.3 & 59.6 & 59.2 & 54.4 & 54.1 \\
Meditron3-8B \cite{meditron3-8b}                        & 82.4 & 77.2 & \textbf{91.1} & \textbf{88.0} & \textbf{60.8} & \textbf{54.2} & \textbf{79.6} & \textbf{72.5} & \textbf{79.8} & \textbf{70.9} \\
\hline
GPT-4.1-mini \cite{openai_gpt4_1_mini} & 
80.2 & 78.0 &
95.5 & 94.5 &
42.3 & 40.8 &
83.3 & 80.1 &
78.8 & 75.6 \\
GPT-4o \cite{openai_gpt4o}&
77.5 & 76.1 &
97.5$^{\dagger}$ & 96.9$^{\dagger}$ &
64.5$^{\dagger}$ & 61.1$^{\dagger}$ &
94.5$^{\dagger}$ & 93.1$^{\dagger}$ &
81.1$^{\dagger}$ & 77.7$^{\dagger}$ \\
\bottomrule
\end{tabular}
\caption{Accuracy and Macro F1(\%) for Language Models across Prompt Templates on \textsc{HealthContradict}.}
\label{tab:template5_comparison}
\end{table*}

The smaller language models, \textsc{LLaMA-3.2-1B-Instruct} and \textsc{BioMed-Llama-3.2-1B}, have the lowest performance, with only 38.3\% and 33.8\% accuracy when using their parametric knowledge in Prompt NC. 
When these models are provided with correct context in Prompt CC, the accuracy drops 9.7 percentage points for the general model ($p < 0.001$), with predictions “yes” most of the time (high recall at 1), whereas for the biomedical model, the accuracy increases by 14.5 percentage points ($p < 0.001$). 
In Prompt IC, the biomedical model \textsc{BioMed-Llama-3.2-1B} can resist the incorrect context with an accuracy 9.2 percentage points higher than the general model \textsc{LLaMA-3.2-1B-Instruct} ($p < 0.001$). 
The biomedical model benefits from a later position of the correct document in conflicting contexts, with a 5.9 percentage points difference between Prompt CIC and ICC ($p = 0.003$), and shows a strong ability to exploit long biomedical context when provided with conflicting context.

The larger language models show better performance. When using only parametric knowledge, \textsc{Meditron3-8B} outperforms \textsc{Llama-3.1-8B-Instruct} by 4.5 percentage points ($p = 0.006$), whereas \textsc{Meditron3-Qwen2.5-7B} shows 0.5 percentage points difference from \textsc{Qwen2.5-7B} ($p = 0.551$). 
It is hard to tell whether biomedical domain fine-tuning has improved the models' capacity under this setting. 
However, when correct biomedical context is introduced, \textsc{Meditron3-8B} outperforms \textsc{LLaMA-3.1-8B-Instruct} by 21.1 percentage points ($p < 0.001$), and \textsc{Meditron3-Qwen2.5-7B} also outperforms \textsc{Qwen2.5-7B} by 10.9 percentage points ($p < 0.001$). 
These results suggest that the fine-tuned biomedical models can exploit correct context much better than their general domain counterparts. 

Although introducing incorrect biomedical context reduces performance across all models, \textsc{Meditron3-8B} remains more robust and achieves 60.8\% accuracy, which is 23.0 percentage points higher than that of \textsc{LLaMA-3.1-8B-Instruct} ($p < 0.001$). In contrast, \textsc{Meditron3-Qwen2.5-7B} does not show the same resistance. These results suggest the instruction-fine-tuned biomedical model \textsc{Meditron3-8B}, exhibits greater robustness under misleading context compared to non-instruction-fine-tuned biomedical model \textsc{Meditron3-Qwen2.5-7B}.
However, \textsc{Meditron3-Qwen2.5-7B} performs 7.0 percentage points better under Prompt CIC than Prompt ICC ($p < 0.001$), indicating that it benefits from the earlier position of the correct document in conflicting contexts.

We show the performance of commercial LLMs. Both \textsc{GPT-4.1-mini} and \textsc{GPT-4o} show the same performance patterns as the open-source models. They are positively influenced by correct contextual knowledge and negatively influenced by incorrect contextual knowledge. When both correct and incorrect contextual knowledge are present, \textsc{GPT-4.1-mini} and \textsc{GPT-4o} lean towards better performance when the correct contextual knowledge appears before the incorrect contextual knowledge. Compared to open-source biomedical models, when the models utilize parametric knowledge, \textsc{Meditron3-Qwen2.5-7B} can outperform both \textsc{GPT-4.1-mini} and \textsc{GPT-4o}. Moreover, \textsc{Meditron3-8B} can outperform \textsc{GPT-4.1-mini} in resisting incorrect contextual knowledge. However, \textsc{GPT-4o} performs the best throughout the contextual prompts. 

We next focus on the two best-performing open-source models, \textsc{Meditron3-8B} and its general-domain counterpart \textsc{Llama3.1-8B-Instruct}, comparing their error types and contextual reasoning abilities, and illustrating the findings with a case study.

\subsubsection{Error Types}

\begin{figure}[!h]
    \centering
    \includegraphics[width=0.5\linewidth]{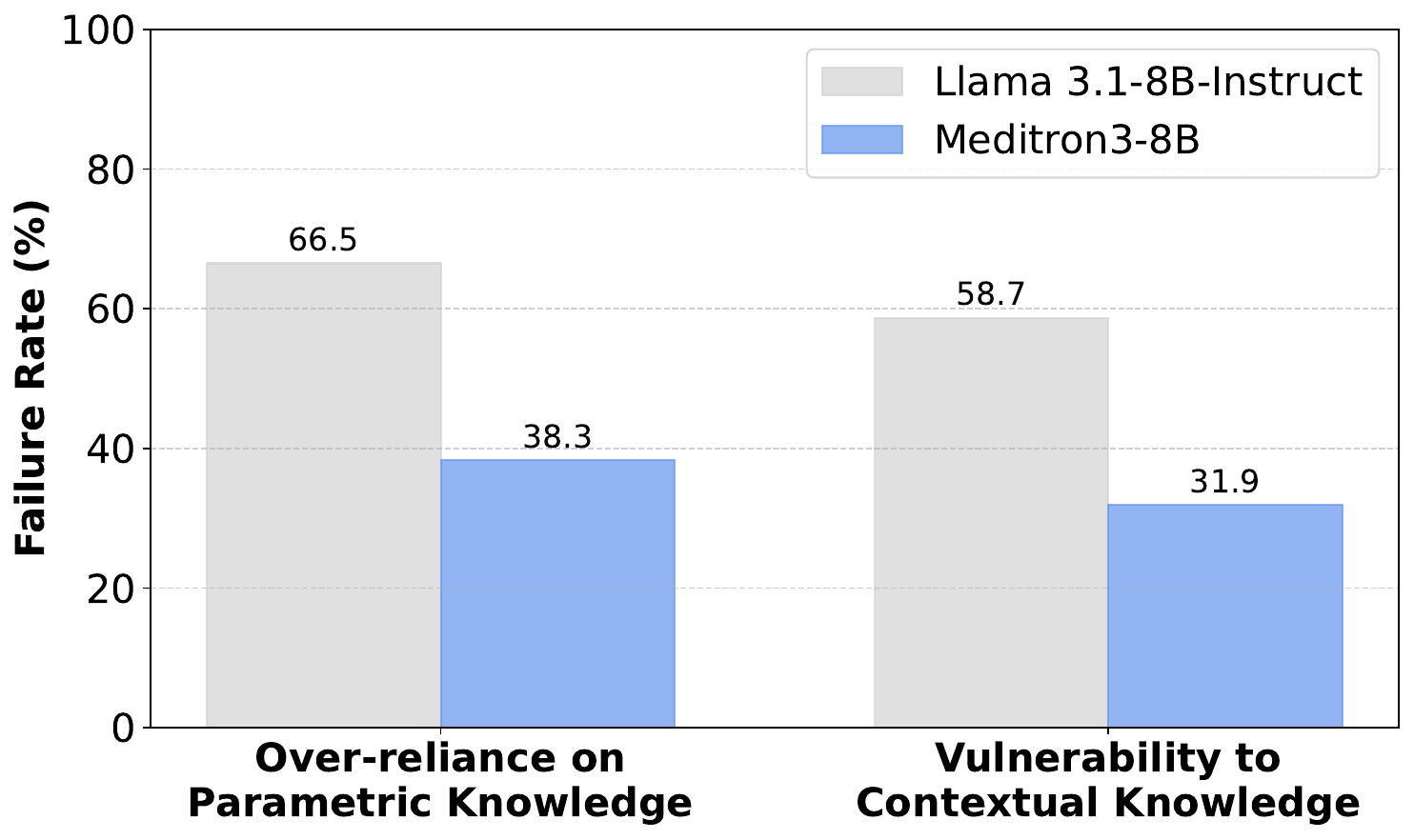}
    \caption{Failure Rates across Error Types}
    \label{fig:error-type}
\end{figure}

We analyze two failure modes with conditional failure rates detailed in the Methods section. The first, over-reliance on parametric knowledge, occurs when the model fails to update an incorrect answer even when provided with correct context (Eq.\ref{eq5}). The second, vulnerability to contextual knowledge, occurs when the model initially answers correctly but changes to an incorrect answer after being shown misleading context (Eq. \ref{eq6}). As shown in Figure \ref{fig:error-type}, the biomedical model Meditron3-8B exhibits over-reliance on parametric knowledge in 38.3\% of cases (vs. 66.5\% for its base model, Llama 3.1-8B-Instruct) and is misled by incorrect context in 31.9\% of cases (vs. 58.7\% for its base model). Overall, the biomedical model makes fewer errors, and language models are more likely to fail due to over-reliance on parametric knowledge than due to misleading context.

\begin{figure*}[!h]
    \centering

    % Bar chart subfigure
    \begin{subfigure}[b]{0.60\textwidth}
        \centering
        \includegraphics[width=\linewidth]{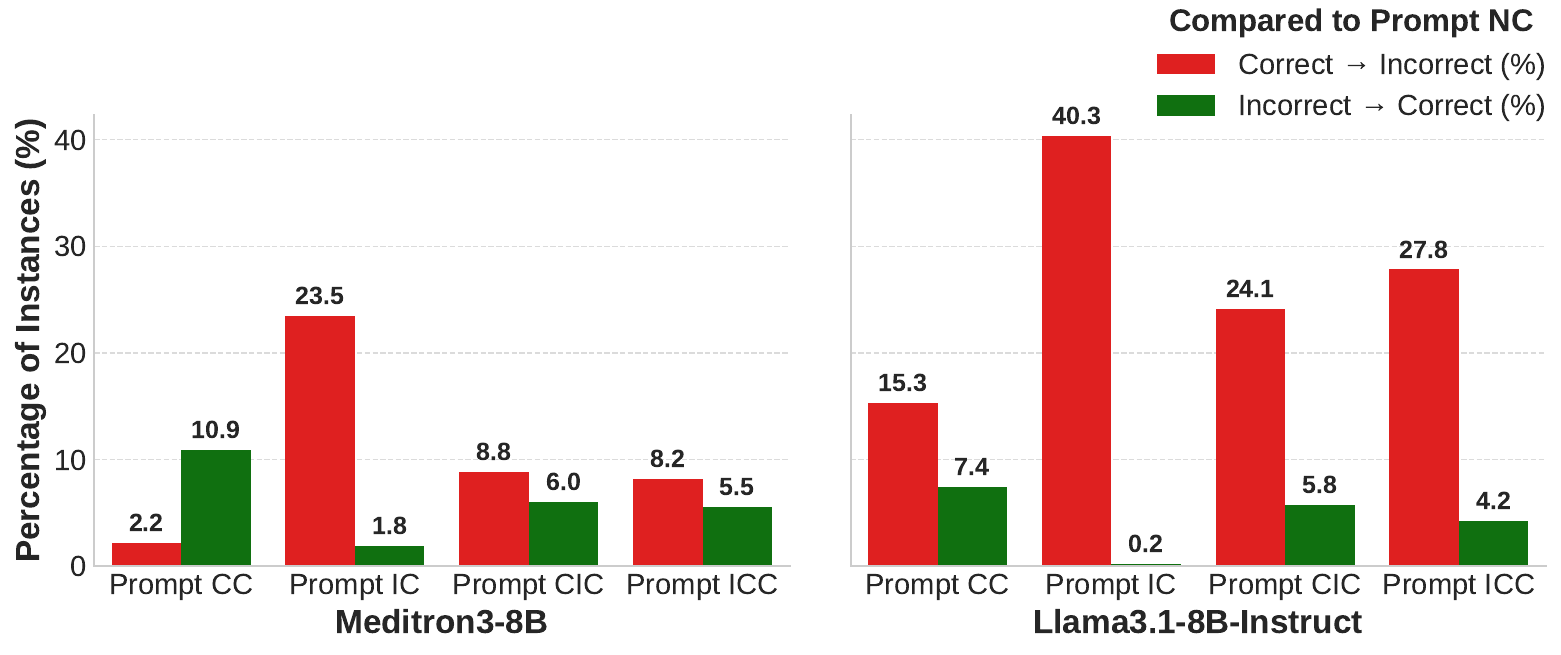}
         \caption{Impact of Context on Model Predictions.}
        \label{fig:accuracy-switch-comparison}
    \end{subfigure}
    \hspace{0.03\textwidth}
    
    % Heatmap subfigure
    \begin{subfigure}[b]{0.60\textwidth}
        \centering
        \includegraphics[width=\linewidth]{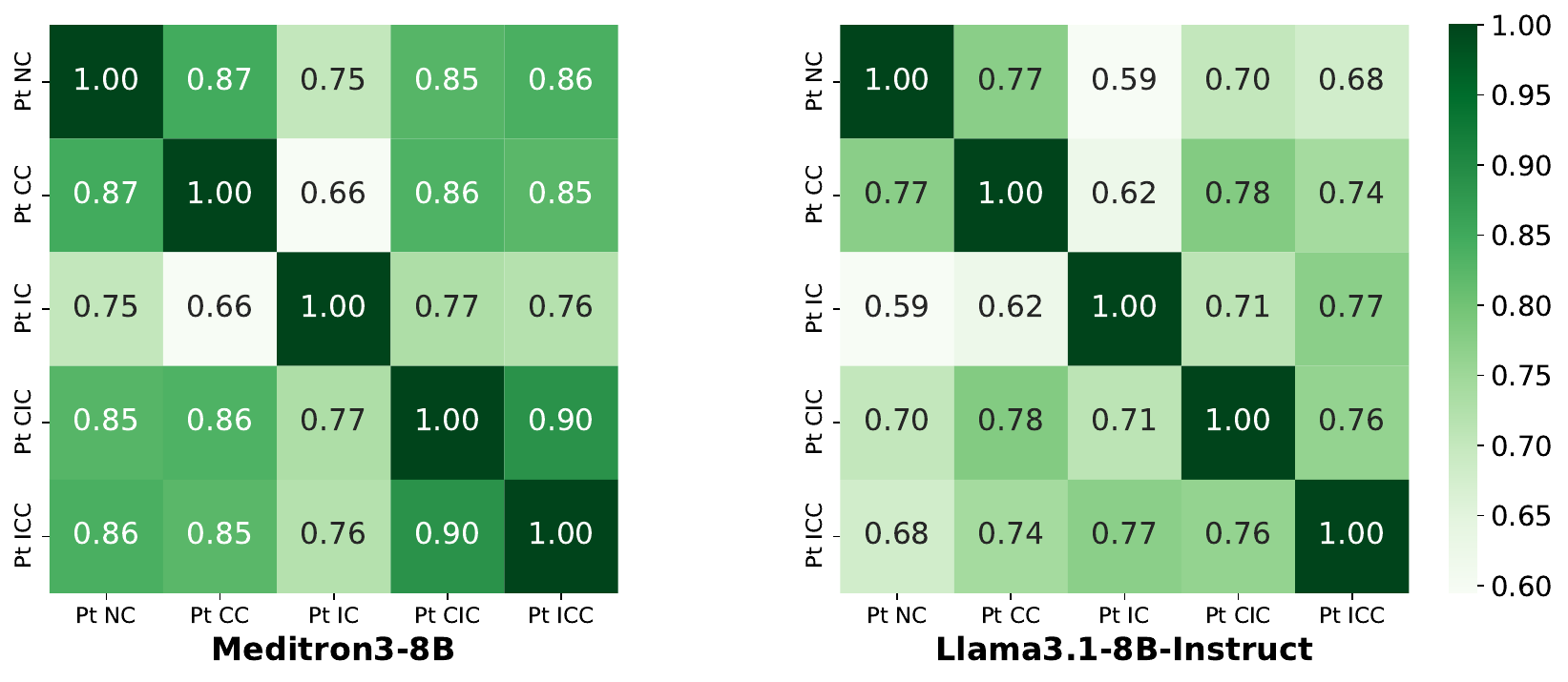}
        \caption{Prediction Agreement between Prompt Templates}
        \label{fig:prompt-agreement-comparison}
    \end{subfigure}

    \caption{Consistency of Predictions across Prompt Templates}
    \label{fig:prompt-sensitivity-overview}
\end{figure*}

To further characterize these error patterns, we analyze models' predictions across Prompt Templates. Figure~\ref{fig:accuracy-switch-comparison} shows the percentage of the context-induced answers. Using the correct biomedical context, \textsc{Meditron3-8B} switches to the correct answer on 10.9\% instances while \textsc{Llama3.1-8B-Instruct} is very confused and switches to incorrect answers on 15.3\% instances. When the incorrect context is introduced, \textsc{Meditron3-8B} shows a much lower rate of switching to incorrect answers (23.5\%) compared to \textsc{Llama3.1-8B-Instruct} (40.3\%). Meanwhile, when they encounter contradictory context, \textsc{Meditron3-8B} switches to correct answers in 8.8\% of cases under Prompt CIC and 8.2\% under Prompt ICC, and switches to incorrect answers in 6.0\% (Prompt CIC) and 5.5\% (Prompt ICC). \textsc{Llama3.1-8B-Instruct} shows a much higher rate of confusion which switches to incorrect answers in 24.1\% (Prompt CIC) and 27.8\% (Prompt ICC) of cases.

We calculate the agreement of predictions across different templates to complement the error-type analysis. As shown in Figure~\ref{fig:prompt-agreement-comparison}, for \textsc{Meditron3-8B}, the most significant difference is between Prompt CC and IC, with an agreement of 0.66. Prompt CIC and ICC show the closest agreement of 0.90. For \textsc{Llama3.1-8B-Instruct}, the most significant difference is between Prompt CC and Prompt IC, at 0.62. Moreover, Prompt CC and Prompt CIC show the closest agreement of 0.78. 

\subsubsection{Context Reasoning}

\begin{figure}[!h]
    \centering
    \includegraphics[width=0.5\linewidth]{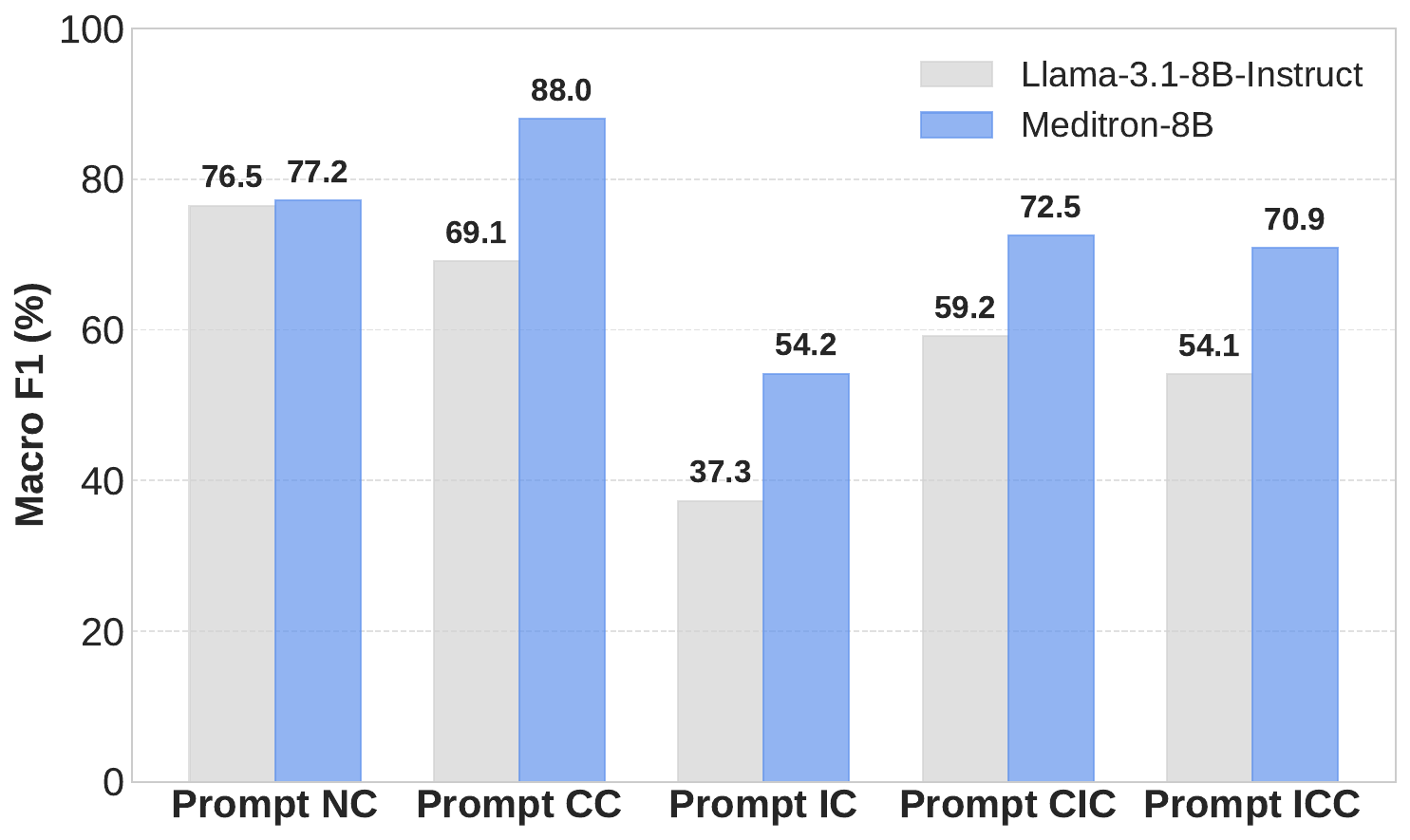}
    \caption{Macro F1(\%) of \textsc{Meditron3-8B} and \textsc{Llama-3.1-8B-Instruct} across Prompt Templates.}
    \label{fig:f1_comparison}
\end{figure}

As shown in Figure~\ref{fig:f1_comparison}, when there is no context, both models answer health questions with their parametric knowledge, and \textsc{Meditron3-8B} has 0.7 percentage points performance improvement on the Macro F1 compared to \textsc{Llama-3.1-8B-Instruct}. However, with the context of a correct document, \textsc{Meditron3-8B} outperforms by 18.9 percentage points. Although the incorrect context induces both models, \textsc{Meditron3-8B} can resist the incorrect context and differentiate from \textsc{Llama-3.1-8B-Instruct} by 16.9 percentage points. When encountering contradictory contextual knowledge, \textsc{Meditron3-8B} also outperforms \textsc{Llama-3.1-8B-Instruct} by 13.3 percentage points and 16.8 percentage points, respectively. The comparative analysis indicates that models fine-tuned for the biomedical domain can exploit correct while resisting incorrect biomedical context. \textsc{HealthContradict} can differentiate models' capacity for long-context biomedical reasoning, particularly in generating factual answers when presented with conflicting biomedical contextual knowledge.

\begin{figure*}[!ht]
    \centering
    \includegraphics[width=\linewidth]{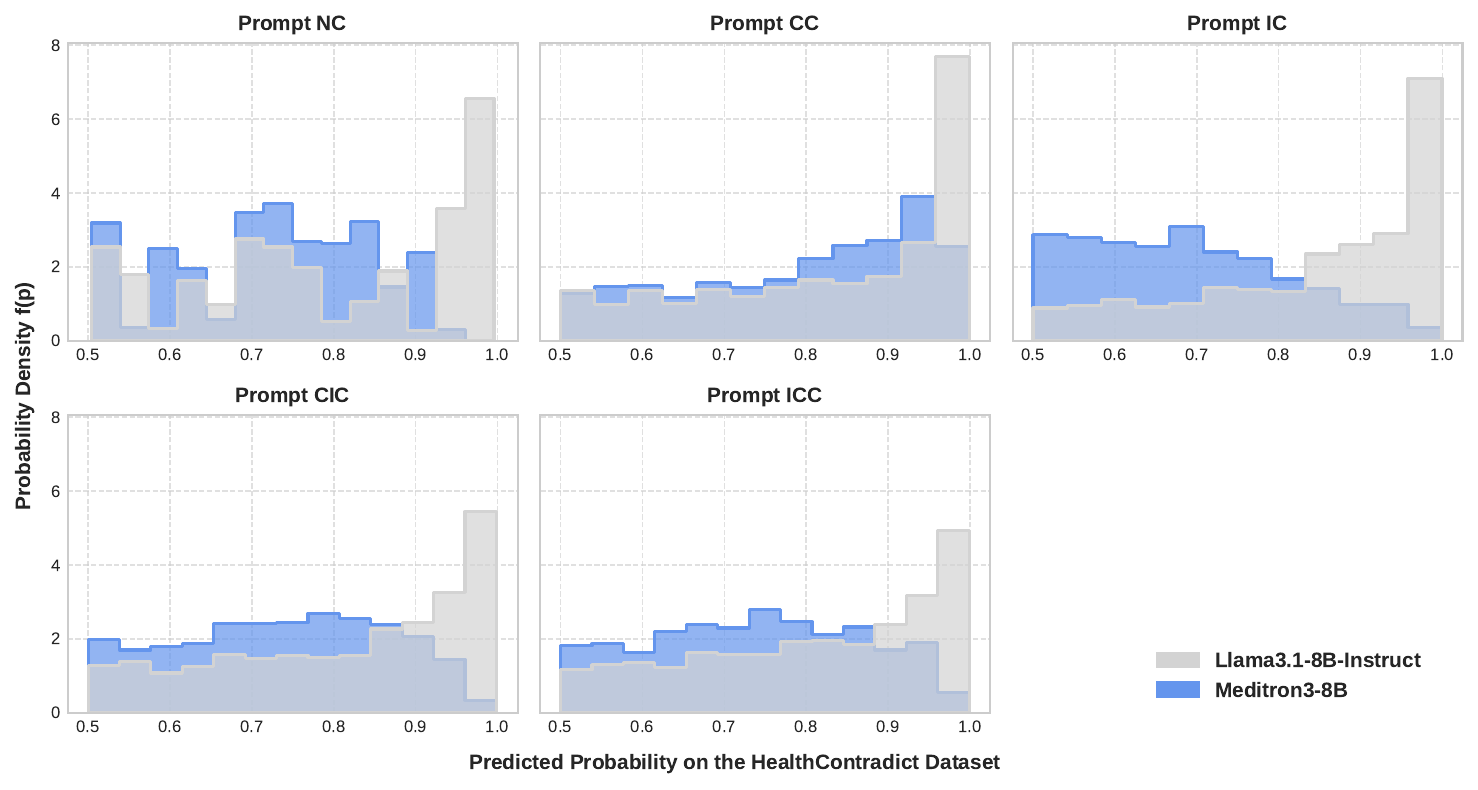}
    \caption{Model Probability Distributions across Prompt Templates.}
    \label{fig:probability-prompts}
\end{figure*}

\begin{figure}[!h]
    \centering
    \includegraphics[width=0.8\linewidth]{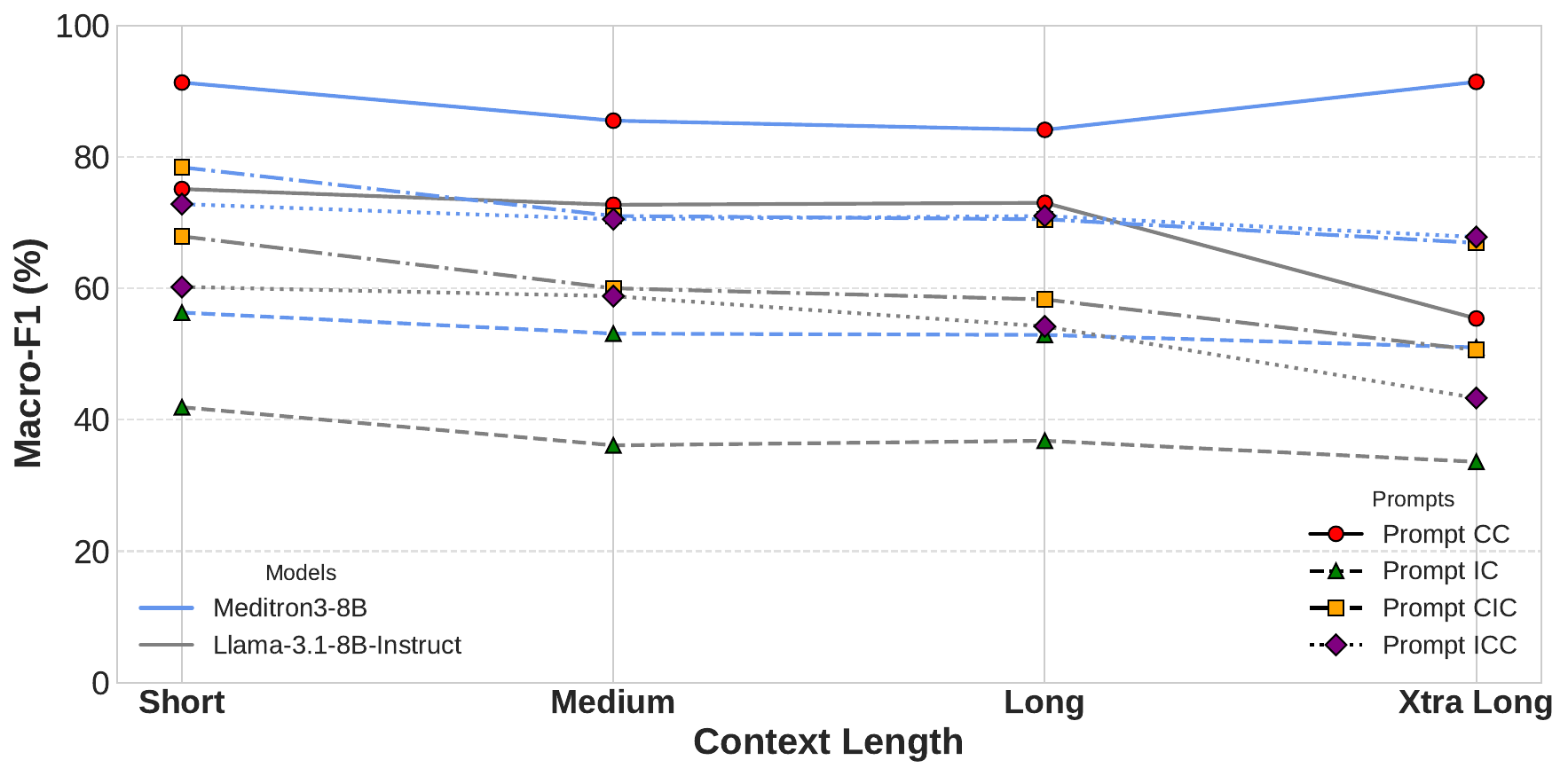}
    \caption{Impact of Input Length on Macro--F1.}
    \label{fig:input_length}
\end{figure}

As shown in Figure~\ref{fig:probability-prompts}, the x-axis represents the predicted probability $\hat{p}_i$ of the model’s answer $\hat{y_i}$ on the \textsc{HealthContradict} dataset, ranging from 0.5 (low probability) to 1.0 (high probability). The predicted probability $\hat{p}_i$ is obtained by extracting the output logits and compute softmax probabilities over the candidate labels \texttt{YES} and \texttt{NO} (Eq.\ref{eq1}, \ref{eq2} and \ref{eq3}). The y-axis represents the estimated probability density $f(p)$ (Eq.\ref{eq4}), which shows how often the model produces predictions at different probability levels. A higher density value indicates that a larger fraction of predicted probabilities are concentrated within that probability range. 

\textsc{LLaMA-3.1-8B-Instruct} exhibits consistently high probability across all templates, with distributions concentrated near 1.0, suggesting high-probability predictions even when the provided context is factually incorrect or contradictory.
In contrast, \textsc{Meditron3-8B} shows an adaptive probability. For Prompt CC, which includes correct biomedical context, \textsc{Meditron3-8B} shows a right shift in its probability scores, indicating increased certainty in its predictions. For Prompt IC, which includes incorrect biomedical context, the probability scores shift left, indicating decreased certainty in its predictions. Furthermore, when presented with conflicting contextual knowledge in Prompt CIC and ICC, \textsc{Meditron3-8B} is with a broader spread of predicted probabilities. The probability distributions indicate that the biomedical domain adapted model modulates its prediction probability based on the factuality of the provided context.

We further examine how the models' contextual reasoning abilities change as context length increases. We partitioned \textsc{HealthContradict} into four groups of equal sample size based on the range of the context length within each prompt template. In Figure \ref{fig:input_length}, both models perform best when the context length is short. But Meditron3‑8B shows robustness compared to its base model on the longer context lengths.

We then show the context reasoning in an interpretable format, a case study for the question -- \textit{``Can cell phones cause cancer?''} -- is illustrated in Table~\ref{tab:prompt-case-study-2}. According to scientific evidence, the factual answer is \texttt{no}. Using parametric knowledge, both \textsc{Meditron3-8B} and \textsc{LLaMA-3.1-8B-Instruct} predict the correct label (\(\hat{y} = \texttt{no}\)) with probability scores of 0.74 and 0.72, respectively. Adding the correct context, \textsc{Meditron3-8B} increases its probability to 0.96, while \textsc{LLaMA-3.1-8B-Instruct} reduces its probability to 0.57. Adding the incorrect context, \textsc{Meditron3-8B} maintaines the correct answer with moderate probability (0.59), whereas \textsc{LLaMA-3.1-8B-Instruct} is misled and predicts \texttt{yes} with high probability (0.87). Adding the contradictory context, both models perform better when the correct information appears later. \textsc{Meditron3-8B} further gives the correct answer with probability scores of 0.62 (Prompt CIC) and 0.86 (Prompt ICC). In contrast, \textsc{LLaMA-3.1-8B-Instruct} fails to identify the correct label in Prompt CIC and only recovers it in Prompt ICC with a lower probability score of 0.58. The case study shows, in an interpretable way, the fine-tuned biomedical language models' capability to better integrate correct contextual knowledge while refusing incorrect knowledge in the answer. 

\begin{table*}[!ht]
\small
\centering
\begin{tabular}{@{}c p{6cm} >{\centering\arraybackslash}m{3cm} >{\centering\arraybackslash}m{3.6cm}@{}}
\toprule
 & \textbf{Prompt Template} & \textsc{Meditron3-8B} & \textsc{Llama-3.1-8B-Instruct}\\
\midrule
\textbf{1} & ...\textbf{Question:} \colorbox{SkyBlue}{Can cell phones cause cancer?} 
 & \raisebox{-.5\height}{\adjustbox{valign=m}{\includegraphics[width=2.5cm]{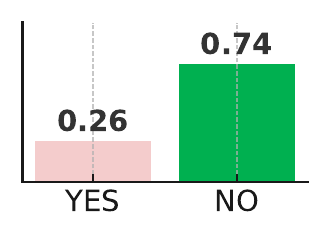}}}
 & \raisebox{-.5\height}{\adjustbox{valign=m}{\includegraphics[width=2.5cm]{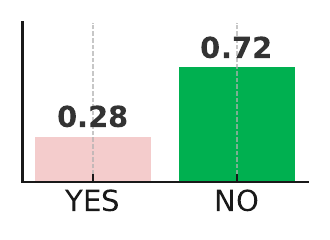}}}\\
\addlinespace
\textbf{2} & ... \textbf{Context:} ... While cell phone use jumped dramatically from 1974 to 2003, the period which the study covers, overall \colorbox{YellowGreen}{brain cancer trends} in the population \colorbox{YellowGreen}{didn't follow suit}...
& \raisebox{-.5\height}{\adjustbox{valign=m}{\includegraphics[width=2.5cm]{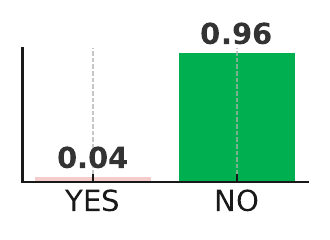}}}
& \raisebox{-.5\height}{\adjustbox{valign=m}{\includegraphics[width=2.5cm]{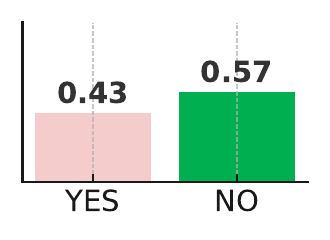}}}\\
\addlinespace
\textbf{3} & ... \textbf{Context:} ...WHO/IARC has classified radiofrequency electromagnetic fields as possibly carcinogenic to humans (group 2B), based on \colorbox{Peach}{an increased risk for glioma}...
& \raisebox{-.5\height}{\adjustbox{valign=m}{\includegraphics[width=2.5cm]{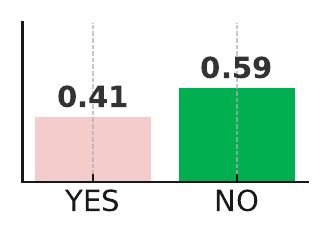}}}
& \raisebox{-.5\height}{\adjustbox{valign=m}{\includegraphics[width=2.5cm]{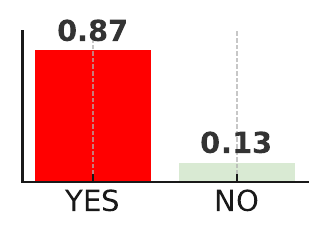}}}\\
\addlinespace
\textbf{4} & ...\textbf{Context:} ...\colorbox{YellowGreen}{brain cancer trends} in the population \colorbox{YellowGreen}{didn't follow suit}...
...\colorbox{Peach}{an increased risk for glioma}...
& \raisebox{-.5\height}{\adjustbox{valign=m}{\includegraphics[width=2.5cm]{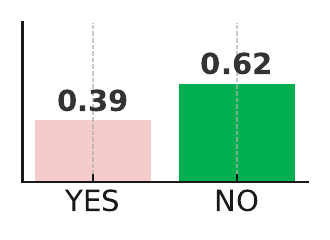}}}
& \raisebox{-.5\height}{\adjustbox{valign=m}{\includegraphics[width=2.5cm]{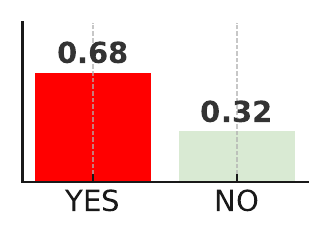}}}\\
\addlinespace
\textbf{5} & ...\textbf{Context:} ...\colorbox{Peach}{an increased risk for glioma}...\newline
... \colorbox{YellowGreen}{brain cancer trends} in the population \colorbox{YellowGreen}{didn't follow suit}... 
& \raisebox{-.5\height}{\adjustbox{valign=m}{\includegraphics[width=2.5cm]{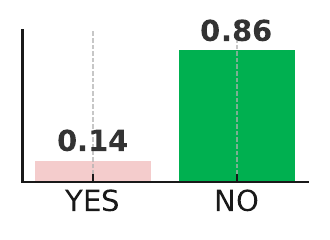}}}
& \raisebox{-.5\height}{\adjustbox{valign=m}{\includegraphics[width=2.5cm]{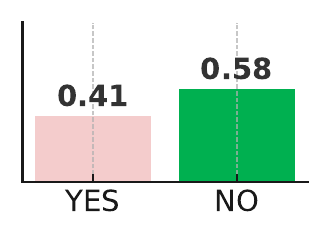}}}\\
\bottomrule
\end{tabular}
\caption{Model probability Scores for \textsc{Meditron3-8B} and \textsc{Llama-3.1-8B-Instruct} on the Question “Can cell phones cause cancer?”}
\label{tab:prompt-case-study-2}
\end{table*}

\section{Discussion}

The \textsc{HealthContradict} benchmark evaluates language models' robustness when encountering biomedical knowledge conflicts. Unlike conventional QA tasks, our use case–oriented design assumes that users may provide incorrect context and often cannot verify the factual accuracy of retrieved documents. The Prompt IC setting specifically tests whether the model follows the instruction to use context or defaults to parametric knowledge. Our findings show that biomedical domain-adapted language models can outperform general-purpose models in biomedical applications when users provide incomplete or incorrect information and lack the expertise to verify factual accuracy. This work shows promising directions for developing safer and more inclusive digital health systems. However, the current accuracy is not yet sufficient for communities with limited medical expertise to rely on. The results demonstrate the critical role of human experts and accountability, especially in settings where misinformation can have serious consequences.

To ensure reproducibility on the resource-limited healthcare environments, we focused on language models between 1B and 8B parameters. While larger models such as ChatGPT are of considerable interest, their limited transparency, fine-tuning restrictions, and unresolved privacy concerns  limit them for clinical deployment.

This study has limitations. We were unable to perform experiments on even larger language models because of resource constraints. Our analysis did not fully capture the reasoning processes behind model predictions in human language due to  models generated justifications may not reflect the actual decision process \cite{chen2025reasoningmodelsdontsay} and limited access to human evaluators. Our prompt templates are grounded in prior work \cite{HouEtAl2024_WikiContradict}. However, variables exist between the Prompt NC (using only parametric knowledge) and the other four prompts (different formats of contextual knowledge), which could influence model performance. While the benchmark is constrained by the scope of the TREC resources and may not fully capture evolving public health issues, it approximates real-world scenarios as closely as possible. We relied on the existing TREC annotations and did not conduct clinical risk analysis due to limited access to clinicians. The question topics were selected by TREC organizers. The documents were judged by NIST assessors following the official guidelines \cite{trec_health_misinfo}. NIST does not disclose the number of assessors used in the TREC Health Misinformation track. 

In conclusion, we present a new benchmark \textsc{HealthContradict}, which uses interpretable quantitative metrics to evaluate language models' ability to reason over long and conflicting biomedical contexts. Compared to state-of-the-art medical QA benchmarks, \textsc{HealthContradict} better captures the difference of models’ performance. In our evaluation, language models adapted to the biomedical domain  show improved ability to (i) leverage correct contextual information, (ii) resist incorrect contextual information, and (iii) decipher between conflicting contextual information.

\section{Methods}

\subsection{Data Collection}
Our benchmark uses expert-annotated questions and documents from the TREC Health Misinformation Track 2019, 2021, and 2022 \cite{Abualsaud2020OverviewOT, clarke2021overview, clarke2022overview}. The selected tracks focused on questions of people seeking health advice online. Each question consists of a health treatment and a health issue. The document pools are ClueWeb12-B13 \cite{clueweb12} for the 2019 track and a no-clean version of the C4 dataset \cite{allenai_c4} for the 2021 and 2022 tracks. Experts annotated each question with a factual answer supported by a separate credible webpage referring to relevant scientific evidence. Web documents were retrieved using either manual or automated retrieval methods and were annotated by experts based on assessments of relevance, efficacy, and credibility. We excluded the 2020 Track as it is incompatible with settings of the other years, focusing on COVID-19 and uses CommonCrawl News (January–April 2020) \cite{} as the document pool.

To the best of our knowledge, the TREC Health Misinformation tracks remain the only publicly available resource that provides (1) expert-curated health questions with ground-truth answers and supporting scientific evidence, (2) document pools containing both supporting and refuting evidence, verdicted by experts. These components are essential for constructing health questions with a ground-truth answer, using pairs of contradictory documents to evaluate biomedical knowledge conflicts in language models.

In \textsc{HealthContradict}, we focus on relevant and credible documents that indicate the efficacy of a document in supporting the answer to the query. We unified labels by mapping the 2019 annotations ``effective''/``ineffective'' and the 2021 annotations ``supportive''/``dissuades'' to the ``yes''/``no'' format, consistent with the 2022 labels. We also consider that if two documents for the same question have opposite stances -- one \texttt{yes}, one \texttt{no} -- they are considered a contradictory pair. Moreover, we define a document as correct if its stance aligns with the scientific evidence, and incorrect if its stance contradicts the scientific evidence.

The original collections include 130 expert-annotated questions (50 in 2019, 35 in 2021, and 45 in 2022). Questions with yes/no stance annotations, as well as those that have both supporting and refuting documents, are included. Applying this criterion reduces the set to 110 questions. We pair supporting and refuting documents into contradiction pairs, ensuring that each document appears only once because some documents may be associated with multiple questions. We exclude questions that cannot form at least one contradiction pair due to insufficient unique documents. This results in a final set of 81 questions and 920 pairs of contradictory documents.

\subsection {Answer Prediction}

Each prompt \(x_i\) is tokenized and processed. We extract the output logits  \(z\) at the final token position and compute softmax probabilities over the candidate labels \texttt{YES} and \texttt{NO}:
\begin{equation}
\label{eq1}
\begin{aligned}
p(y_i \mid x_i) = \frac{e^{z_{y_i}}}{e^{z_{\texttt{YES}}} + e^{z_{\texttt{NO}}}} \quad \text{for } y_i \in \{\texttt{YES}, \texttt{NO}\}.
\end{aligned}
\end{equation}
The predicted label \(\hat y_i\) with the highest probability for the model’s prediction is defined as:
\begin{equation}
\label{eq2}
\hat{y_i} = \mathop{\arg\max}\limits_{y \in \{\texttt{YES}, \texttt{NO}\}} p(y_i \mid x_i).
\end{equation}
We denote by $\hat{p}_i$ the corresponding predicted probability value: 
\begin{equation}
\label{eq3}
\hat{p}_i = p(\hat{y_i} \mid x_i).
\end{equation}

To visualize the overall distribution of the predicted probabilities, we computed their empirical probability density using normalized histograms. For each model and prompt template, the predicted probabilities $\hat{p}_i$ were grouped into bins width~$\Delta p$, and the density in each bin center~$p_j$ was defined as
\begin{equation}
\label{eq4}
f(p_j) = \frac{n_j}{N \, \Delta p},
\end{equation}
where $n_j$ denotes the number of predictions within bin~$j$, $N$ is the total number of predictions, and $\Delta p$ is the width of the bin. The densities were normalized so that $\sum_j f(p_j)\Delta p = 1$.

\subsection{Failure Modes}

We defined two failure modes to further analyze the model's performance:

\textbf{Over-reliance on parametric knowledge (OR)} occurs when the model fails to update an incorrect answer even after being provided with correct context (i.e., comparing \textit{Prompt NC} and \textit{Prompt CC}). OR shows how much the model depends on its parametric knowledge and how resistant it is to incorporating correct contextual knowledge. To compute it, we denote by $I_\text{NC}$ the set of instances with incorrect model answers for \textit{Prompt NC}, $I_\text{CC}$ the the set of instances with incorrect model answers for \textit{Prompt CC}, and $N$ the number of instances. The OR rate is then defined as:

\begin{equation}
\label{eq5}
P_{\mathrm{OR}}
= \mathbb{P}\!\left(I_{\mathrm{CC}} \mid I_{\mathrm{NC}}\right)
= \frac{\mathbb{P}\!\left(I_{\mathrm{CC}} \cap I_{\mathrm{NC}}\right)}
        {\mathbb{P}\!\left(I_{\mathrm{NC}}\right)}
= \frac{N_{I_{\mathrm{CC}}\cap I_{\mathrm{NC}}}}{N_{I_{\mathrm{NC}}}}.
\end{equation}

\textbf{Vulnerability to misleading context (VM)} occurs when the model initially provides a correct answer using parametric knowledge but changes to an incorrect answer after being provided with an incorrect context (i.e., comparing \textit{Prompt NC} and \textit{Prompt IC}). VM shows how vulnerable the model is to be misled by incorrect contextual knowledge. To compute it, we denote by $C_{\mathrm{NC}}$ the set of instances with correct model answers for \textit{Prompt NC}, $I_{\mathrm{IC}}$ the set of instances with incorrect model answers for \textit{Prompt IC}, and $N$ the number of instances. Then, the VM rate is defined as:
\begin{equation}
\label{eq6}
P_{\mathrm{VM}}
= \mathbb{P}\!\left(I_{\mathrm{IC}} \mid C_{\mathrm{NC}}\right)
= \frac{\mathbb{P}\!\left(I_{\mathrm{IC}} \cap C_{\mathrm{NC}}\right)}
        {\mathbb{P}\!\left(C_{\mathrm{NC}}\right)}
= \frac{N_{I_{\mathrm{IC}}\cap C_{\mathrm{NC}}}}{N_{C_{\mathrm{NC}}}} .
\end{equation}

\subsection {Evaluation Metrics}

We report model performance using two standard metrics: Accuracy and Macro F1. Accuracy measures the proportion of questions for which the model predicts the correct answer, and is defined as:
\begin{equation}
\label{acc}
\text{Accuracy} = \frac{1}{N} \sum_{i=1}^{N} \mathbf{1}\!\left(\hat{y}_{i} = y_{i}\right),
\end{equation}
where \(N\) is the total number of instances, \(y_i\) is the ground-truth label, and \(\hat{y}_i\) is the predicted label. Precision, Recall and F1-score are given by:

\begin{equation}
\label{f1}
\text{Precision} = 
\frac{\mathrm{TP}}{\mathrm{TP} + \mathrm{FP}}, 
\qquad
\text{Recall} = 
\frac{\mathrm{TP}}{\mathrm{TP} + \mathrm{FN}},
\qquad
\text{F1} = 
\frac{2 \cdot \text{Precision} \cdot \text{Recall}}
{\text{Precision} + \text{Recall}},
\end{equation}
where TP, FP, and FN denote true positives, false positives, and false negatives for each class. Macro-F1 is then obtained by averaging F1-score across the two classes:
\begin{equation}
\label{mf1}
\text{Macro\text{-}F1} = 
\frac{1}{2} \left( \text{F1}_{\text{YES}} + \text{F1}_{\text{NO}} \right).
\end{equation}

\subsection{Statistical Analysis}

The statistical significance was assessed using McNemar’s test, which is appropriate for paired binary predictions. For each comparison, we constructed a 2×2 contingency table from instance-level correctness and reported the chi‑square statistic and two‑sided p‑values with continuity correction.

\subsection{Experiments and Hardware}

During the evaluation, we use pre-trained language models and tokenizers from HuggingFace's transformers library. Models are loaded with FP16 precision and Flash Attention 2. All experiments are run on a single NVIDIA A100 40GB GPU.

\section*{Author Contributions Statement}

B.Z., A.B. and D.T. contributed to the concept,  B.Z. conducted the experiments, analysed the results and drafted the manuscript. B.Z., A.B., R.Y., N.L. and D.T. contributed to the review and revision of the paper. 

\section*{Additional Information}

The complete code and dataset are available on \href{https://github.com/tinaboya/HealthContradict}{github.com/tinaboya/HealthContradict}.

\bibliography{ref}

\appendix

\renewcommand{\thetable}{A\arabic{table}}
\setcounter{table}{0}

\clearpage

\section{Evaluation Statistics}
\subsection{\textsc{HealthContradict} Dataset}

\begin{table}[!h]
    \centering
    \begin{tabular}{l r}
    \toprule
    \textbf{Statistic} & \textbf{Value} \\
    \midrule
    Instances                    & 920 \\
    Questions                    & 81 \\
    Documents                    & 1,840 \\
    Max doc length (words)       & 30,444 \\
    Min doc length (words)       & 23 \\
    Avg doc length (words)       & 2,347 \\
    Earliest date                & 2012-02-10 \\
    Latest date                  & 2019-04-26 \\
    Unique domains               & 1,403 \\
    Avg URL path length          & 6 \\
    \bottomrule
    \end{tabular}
    \caption{Statistics of \textsc{HealthContradict} dataset.}
    \label{tab:dataset-stats}
\end{table}

\begin{table}[!h]
    \centering
    \begin{tabular}{@{}l r@{}}
    \toprule
    \textbf{Domain} & \textbf{Count} \\
    \midrule
    \texttt{www.emfnews.org}                  & 20 \\
    \texttt{www.drbriffa.com}                 & 20 \\
    \texttt{www.healthline.com}               & 15 \\
    \texttt{cellphoneradiationtoday.com}      & 11 \\
    \texttt{www.quackometer.net}              & 10 \\
    \texttt{www.verywellhealth.com}           & 9 \\
    \texttt{www.medicalnewstoday.com}         & 9 \\
    \texttt{www.phantomvibrationsyndrome.com} & 9 \\
    \texttt{www.webmd.com}                    & 8 \\
    \texttt{www.livestrong.com}               & 8 \\
    \bottomrule
    \end{tabular}
    \caption{Top 10 most frequent domains in the \textsc{HealthContradict} dataset.}
    \label{tab:top-domains}
\end{table}

\iffalse
\subsection{Selected Language Models}
Table~\ref{tab:biomed-models} summarizes the maximum context lengths supported by selected language models, ranging from 32K to 131K tokens.

\begin{table}[!h]
\centering
\begin{tabular}{lr}
\toprule
\textbf{Model} & \textbf{Context Length} \\
\midrule
Llama-3.2-1B-Instruct & 128K \\
Qwen2.5-1.5B-Instruct & 32K \\
Qwen2.5-7B & 131K\\
Meditron3-Qwen2.5-7B & 131K \\
Llama-3.1-8B-Instruct & 128K \\
Meditron3-8B & 128K \\
\bottomrule
\end{tabular}
\caption{Context Lengths of Selected Language Models.}
\label{tab:biomed-models}
\end{table}
\fi

\section{Another Case Study}
\label{ap2}
\begin{table*}[!h]
\centering
\small
\begin{tabular}{@{}c p{6cm} >{\centering\arraybackslash}m{3cm} >{\centering\arraybackslash}m{3.6cm}@{}}
\toprule
 & \textbf{Prompt Description} & \textsc{Meditron3-8B} & \textsc{Llama-3.1-8B-Instruct}\\
\midrule
\textbf{1} & ...\textbf{Question:} \colorbox{SkyBlue}{Can coffee help you lose weight?} 
 & \raisebox{-.5\height}{\adjustbox{valign=m}{\includegraphics[width=3cm]{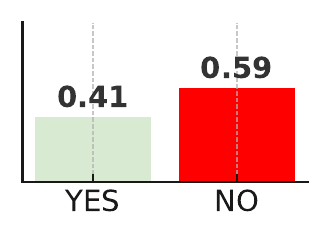}}
} & \raisebox{-.5\height}{\adjustbox{valign=m}{\includegraphics[width=3cm]{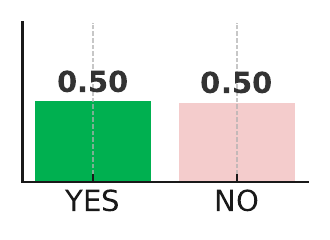}}
}\\
\addlinespace
\textbf{2} & ...\textbf{Context:} ... Green coffee can be taken before or after meals and it is also known to be \colorbox{YellowGreen}{useful for weight loss} ... & \raisebox{-.5\height}{\adjustbox{valign=m}{\includegraphics[width=3cm]{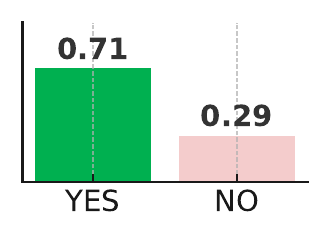}}
} & \raisebox{-.5\height}{\adjustbox{valign=m}{\includegraphics[width=3cm]{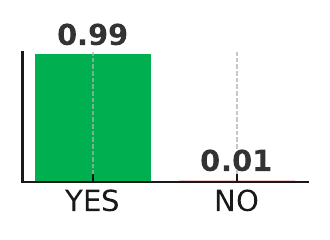}}
}\\
\addlinespace
\textbf{3} & ...\textbf{Context:} ... Does Coffee Can Magically Lose You Weight? \colorbox{Peach}{Of Course Not!} ... & \raisebox{-.5\height}{\adjustbox{valign=m}{\includegraphics[width=3cm]{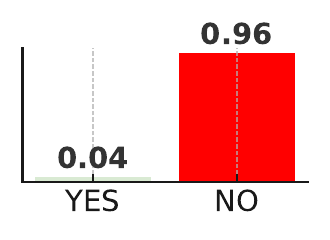}} 
}& \raisebox{-.5\height}{\adjustbox{valign=m}{\includegraphics[width=3cm]{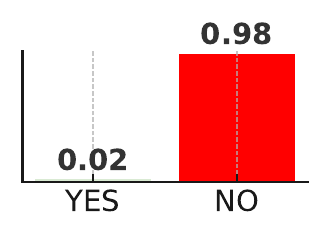}}
}\\
\addlinespace
\textbf{4} & ...\textbf{Context:} ... Green coffee can be taken before or after meals and it is also known to be \colorbox{YellowGreen}{useful for weight loss} ... \newline ... Does Coffee Can Magically Lose You Weight? \colorbox{Peach}{Of Course Not!} ... & \raisebox{-.5\height}{\adjustbox{valign=m}{\includegraphics[width=3cm]{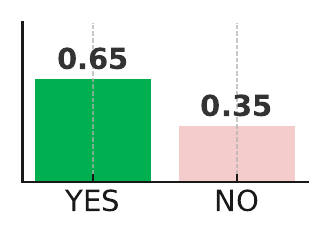}}
}& \raisebox{-.5\height}{\adjustbox{valign=m}{\includegraphics[width=3cm]{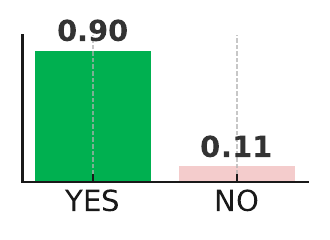}}
}\\
\addlinespace
\textbf{5} & ...\textbf{Context:} ... Does Coffee Can Magically Lose You Weight? \colorbox{Peach}{Of Course Not!} ... \newline ... Green coffee can be taken before or after meals and it is also known to be \colorbox{YellowGreen}{useful for weight loss} ...& \raisebox{-.5\height}{\adjustbox{valign=m}{\includegraphics[width=3cm]{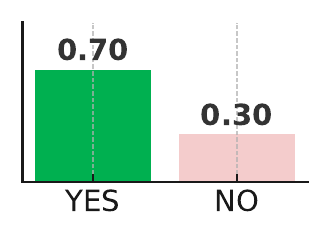}}
}& \raisebox{-.5\height}{\adjustbox{valign=m}{\includegraphics[width=3cm]{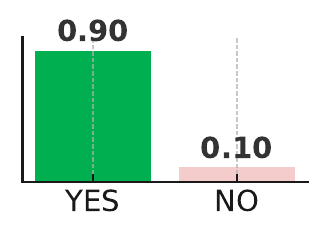}}
}\\
\bottomrule
\end{tabular}
\caption{Model probability Scores for \textsc{Meditron3-8B} and \textsc{Llama-3.1-8B-Instruct} on the Question “Can coffee help you lose weight?”}

\label{tab:prompt-case-study}    
\end{table*}

Table~\ref{tab:prompt-case-study} shows the case study of the same example in Figure~\ref{fig:kg} and Table~\ref{tab:example-instance}. Under the question -- \textit{``Can coffee help you lose weight?''} -- both models are unsure of what to answer with their parametric knowledge. Moreover, \textsc{Meditron3-8B} made the wrong prediction \texttt{NO} with a probability score of 0.59. When adding the correct context, both models give a higher probability in the correct answer, \texttt{YES}. \textsc{Meditron3-8B} could change from an incorrect answer to a correct answer. However, when adding the incorrect context, both models boosted their probability in the incorrect answer \texttt{NO}. Both the finetuned biomedical model and its general domain counterpart fail to resist the incorrect context. When adding contradictory context, both models could identify the correct context and make the right prediction.

\newpage

\section{Disease and Condition Categories of Health Questions in \textsc{HealthContradict}}

\begin{longtable}{p{4cm} p{10cm} p{1.5cm}}
\caption{Health Questions and Their Associated Disease or Condition Categories} \label{tab:full-question-table} \\
\toprule
\textbf{Disease/Condition} & \textbf{Question} & \textbf{TREC ID} \\
\midrule
\endfirsthead
\toprule
\textbf{Disease/Condition} & \textbf{Question} & \textbf{TREC ID} \\
\midrule
\endhead
AIDS & Can HIV be transmitted through sweat? & 183 \\
AIDS & Did AIDS come from chimps? & 181 \\
AIDS & Is male circumcision helpful in reducing heterosexual men's chances of getting HIV? & 12 \\
Alcohol use disorder & Can benzos (benzodiazepines) help with alcohol withdrawal? & 41 \\
Anxiety disorders & Can l-theanine supplements reduce stress and anxiety? & 131 \\
Arthritis & Can you use WD-40 for arthritis? & 155 \\
Arthritis & Do magnetic wrist straps help with arthritis? & 164 \\
Arthritis & Can copper bracelets reduce the pain of arthritis? & 139 \\
Asthma & Can vitamin D supplements improve the management of asthma? & 146 \\
Asthma & Does yoga improve the management of asthma? & 107 \\
Athlete's foot & Can fungal creams treat athlete's foot? & 140 \\
Atopic dermatitis & Can dupixent treat eczema? & 118 \\
Atopic dermatitis & Are probiotics an effective treatment for eczema? & 42 \\
Autism & Are vaccines linked to autism? & 158 \\
Burns & Should I apply ice to a burn? & 105 \\
Cancer & Does deli meat increase your risk of colon cancer? & 190 \\
Cancer & Can cell phones cause cancer? & 154 \\
Cancer & Can cancer be inherited? & 157 \\
Cancer & Can baking soda help to cure cancer? & 159 \\
Cancer & Does selenium help prevent cancer? & 109 \\
Cancer & Is amygdalin or laetrile an effective cancer treatment? & 6 \\
Cavities and tooth decay & Can oil pulling heal cavities? & 192 \\
Common cold & Does Vitamin C prevent colds? & 187 \\
Common cold & Does inhaling steam help treat common cold? & 132 \\
Common warts & Does duct tape work for wart removal? & 104 \\
Croup & Does steam from a shower help croup? & 128 \\
Dementia & Can folic acid help improve cognition and treat dementia? & 103 \\
Depression & Can music therapy help manage depression? & 144 \\
Diabetes & Can fruit juice increase the risk of diabetes? & 163 \\
Diabetes & Can cinnamon help people with diabetes? & 40 \\
Epilepsy & Can vitamins help manage epilepsy? & 44 \\
Erectile dysfunction & Do ACE inhibitors typically cause erectile dysfunction? & 161 \\
Eye problems in adults & Are carrots good for your eyes? & 175 \\
Fever & Is starving a fever effective? & 108 \\
Fever & Is a tepid sponge bath a good way to reduce fever in children? & 102 \\
Foreign object swallowed & Will drinking vinegar dissolve a stuck fish bone? & 137 \\
Hair loss & Can minoxidil treat hair loss? & 129 \\
Hemorrhoids & Does a high fiber diet help with hemorrhoids? & 193 \\
High cholesterol & Can exercise lower cholesterol? & 178 \\
High cholesterol & Can fish oil improve your cholesterol? & 170 \\
Hypertension & Can fermented milk help mitigate high blood pressure? & 117 \\
Iron deficiency anemia & Can eating dates help manage iron deficiency anemia? & 136 \\
Jet lag disorder & Can melatonin be used to reduce jet lag? & 8 \\
Keloid scar & Can applying vitamin E cream remove skin scars? & 114 \\
Kidney stones & Can crystals heal? & 152 \\
Knee Pain & Are squats bad for knees? & 160 \\
Low back pain in adults & Can exercises relieve lower back pain? & 11 \\
Low back pain in adults & Find documents that discuss using antidepressants for helping to manage or relieve lower back pain. & 13 \\
Low back pain in adults & Is lumbar traction an effective treatment for lower back pain? & 38 \\
Low back pain in adults & Can insoles treat back pain? & 47 \\
Migraine & Does Aleve relieve migraine headaches? & 122 \\
Migraine & Can the drug Imitrex (sumatriptan) treat acute migraine attacks? & 120 \\
Mild cognitive impairment & Can statins cause permanent cognitive impairment? & 186 \\
Mosquito bites & Can mosquito bites make you sick? & 156 \\
Muscle cramp & Can magnesium prevent muscle cramps? & 16 \\
Nausea and vomiting & Does ginger help with nausea? & 199 \\
Obesity & Is bariatric surgery effective for obesity? & 50 \\
Opioid Use Disorder & Is morphine addictive? & 162 \\
Osteoarthritis & Will at-home exercises manage hip osteoarthritis pain? & 149 \\
Osteoarthritis & Does Tylenol manage the symptoms of osteoarthritis? & 143 \\
Osteoarthritis & Can collagen supplements cure osteoarthritis? & 153 \\
Other & Are there health benefits to drinking your own urine? & 167 \\
Other & Is pink salt good for you? & 168 \\
Other & Is hydroquinone banned in Europe? & 173 \\
Other & Can chewing gum help lose weight? & 179 \\
Other & Do Himalayan salt lamps have health benefits? & 184 \\
Other & Can coffee help you lose weight? & 188 \\
Other & Does drinking lemon water help with belly fat? & 189 \\
Other & Can grapefruit interfere with medication? & 194 \\
Other & Can vape pens be harmful? & 195 \\
Other & Is wifi harmful for health? & 197 \\
Ovarian cyst & Will taking birth control pills treat an ovarian cyst? & 110 \\
Pneumonia & Can antibiotics be use to treat community acquired pneumonia in children? & 29 \\
Post-extraction bleeding & Do tea bags help to clot blood in pulled teeth? & 151 \\
Pregnancy & Can a woman get pregnant while breastfeeding? & 185 \\
Pregnancy & Can an MRI harm my baby? & 177 \\
Pregnancy & Will taking zinc supplements improve pregnancy? & 111 \\
Rheumatoid arthritis & Is sulfasalazine an effective treatment for rheumatoid arthritis? & 49 \\
Spinal cord injury & Can steroids be used as a treatment for spinal cord injury? & 20 \\
Tick bites & Can I remove a tick by covering it with Vaseline? & 134 \\
Whooping cough & Can antibiotics be used as a treatment for whooping cough (pertussis)? & 28 \\
\bottomrule
\end{longtable}

\end{document}